\documentclass[letterpaper, 10 pt, conference]{ieeeconf}  

\IEEEoverridecommandlockouts                              

\usepackage{times}
\usepackage[pdftex]{graphicx}
\usepackage{subfigure}
\usepackage{amsmath,amssymb,amsopn,amstext,amsfonts}
\usepackage{cancel}
\usepackage[space]{cite}
\usepackage{pdfsync}
\usepackage{balance}
\usepackage{color}
\usepackage{mathtools}
\usepackage{bm}

\usepackage{diagbox}
\usepackage{float}
\usepackage{epstopdf}
\usepackage{pifont}
\usepackage{fixltx2e}
\usepackage{amsmath}
\usepackage{multirow}
\usepackage{url}
\usepackage{verbatim}
\usepackage[linkcolor=black,citecolor=black,urlcolor=black,colorlinks=true]{hyperref}
\usepackage{soul,xcolor}
\usepackage[linesnumbered,ruled,vlined]{algorithm2e}
\usepackage{overpic}
\usepackage{bbding}
\usepackage{amsmath,amssymb}

\graphicspath{{../img/}}
\DeclareGraphicsExtensions{.png,.jpg,.eps,.pdf,.jpeg}

\title{\LARGE \bf
PredRecon: A Prediction-boosted Planning Framework \\ for Fast and High-quality Autonomous Aerial Reconstruction
}

\author{Chen Feng$^{2}$, Haojia Li$^{2}$, Fei Gao$^{3,4}$, Boyu Zhou$^{1,\dag}$, and Shaojie Shen$^{2}$
\thanks{ $^{1}$School of Artificial Intelligence, Sun Yat-Sen University, Zhuhai, China.}
\thanks{$^{2}$Department of Electronic and Computer Engineering, The Hong Kong University of Science and Technology, Hong Kong, China.}
\thanks{$^{3}$State Key Laboratory of Industrail Control Technology, Institute of Cyber-Systems and Control, Zhejiang University, Hangzhou, China. }
\thanks{$^{4}$Huzhou Institute, Zhejiang University, Huzhou, China.}
\thanks{ {\tt\footnotesize $\{$cfengag, hlied, eeshaojie$\}$@ust.hk},}
\thanks{{\tt\footnotesize fgaoaa@zju.edu.cn}, {\tt\footnotesize zhouby23@mail.sysu.edu.cn}}
\thanks{\textbf{$^{\dag}$ Corresponding Author}}
}

\begin{document}

\maketitle
\thispagestyle{empty}
\pagestyle{empty}

\begin{abstract}

Autonomous UAV path planning for 3D reconstruction has been actively studied in various applications for high-quality 3D models.
However, most existing works have adopted \textit{explore-then-exploit}, prior-based or exploration-based strategies, demonstrating inefficiency with repeated flight and low autonomy.
In this paper, we propose \textbf{PredRecon}, a prediction-boosted planning framework that can autonomously generate paths for high 3D reconstruction quality. 
We obtain inspiration from humans can roughly infer the complete construction structure from partial observation.
Hence, we devise a surface prediction module (SPM) to predict the coarse complete surfaces of the target from the current partial reconstruction.
Then, the uncovered surfaces are produced by online volumetric mapping waiting for observation by UAV.
Lastly, a hierarchical planner plans motions for 3D reconstruction, which sequentially finds efficient global coverage paths,
plans local paths for maximizing the performance of \textit{Multi-View Stereo} (MVS), and generates smooth trajectories for image-pose pairs acquisition.
We conduct benchmarks in the realistic simulator, which validates the performance of \textbf{PredRecon} compared with the classical
and \textit{state-of-the-art} methods. The open-source code is released at  \urlstyle{same}\url{https://github.com/HKUST-Aerial-Robotics/PredRecon}.
\end{abstract}

\section{Introduction}
Recently, high-quality 3D reconstruction has been an active topic in various applications including cultural relics digitalization, AR/VR, and structural inspection. 
Due to its high flexibility, the unmanned aerial vehicle (\textbf{UAV}) is ideal to achieve the fast, accurate, and complete 3D reconstruction of the target areas.
To effectively improve reconstruction quality and efficiency, a path planning framework for autonomous aerial reconstruction is essential.

Existing reconstruction planning works \cite{zhang2021continuous, hepp2018plan3d, kuang2020real, zhou2020offsite, song2021view, zhou2021fuel, song2020active} demonstrate 
unsatisfactory efficiency in reconstructing the target areas. First of all, many previous methods \cite{zhang2021continuous, hepp2018plan3d, kuang2020real, zhou2020offsite} 
adopt \textit{explore-then-exploit} strategy which requires two scanning trails, or rely on coarse prior models to obtain the reconstruction paths. 
Such strategies present several drawbacks. 1) Two scanning trails lead to task completion inefficiency.
2) As requiring input prior model, the task cannot be fully automated. 3) They cannot guarantee accurate and complete details of the target areas owing to
planning only based on coarse or prior models, which cannot adjust flight paths in \textit{real-time} based on actual observation. 
Recently, online planning methods requiring a single scanning trail and not relying on prior models have been proposed \cite{song2021view, zhou2021fuel,song2020active}, which
partially resolve the above issues. However, the efficiency is not satisfactory enough, due to the fact that the target areas are previously unknown and significant time
is distributed to explore the unknown regions. Besides, some of them demonstrate a prohibitive computation time, which usually results in undesirable stop-and-go behaviors or even requires communications with external high-end computers.

\begin{figure}[t]
	\begin{center}
      \includegraphics[width=0.88\columnwidth]{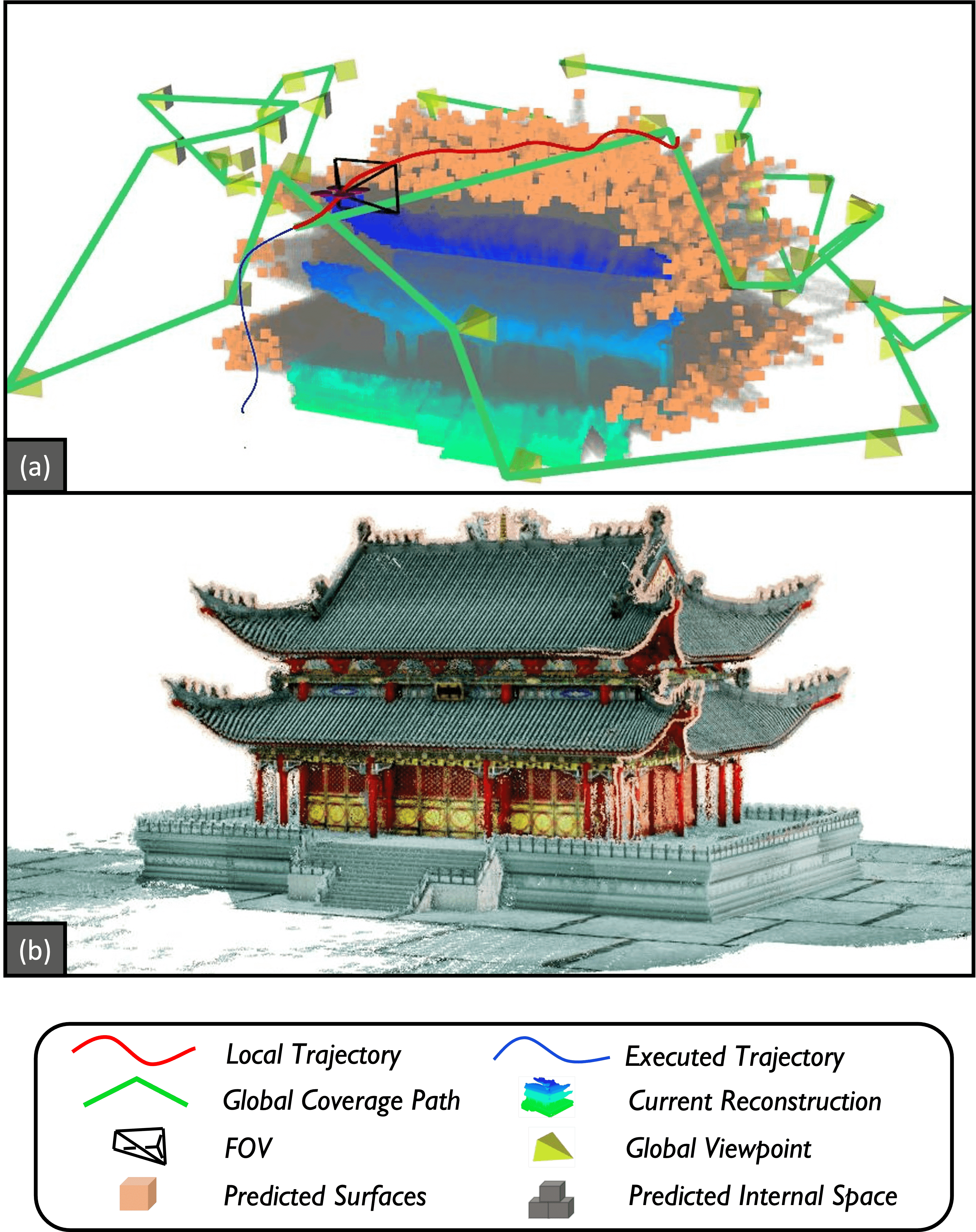}     
      \vspace{-0.5cm}
	\end{center}
   \caption{\label{fig:topo_traj} (a) Illustration of the proposed framework results during executing trajectory for 3D reconstruction, (b) 3D reconstruction result of the above target produced by the proposed framework.
   }
   \vspace{-0.9cm}
\end{figure} 

To address the above issues, we propose \textbf{PredRecon}, a prediction-boosted planning framework that can efficiently reconstruct high-quality 3D models for the target areas in unknown environments with a single flight.
Our method is inspired by the fact that humans can reasonably infer those incomplete structures based on partial observations according to their knowledge and experience. The inferred structures or surfaces enable more purposeful viewpoints generation,
which in turn allows a more efficient global coverage path of the entire target without wasting significant time on exploring unknown space. Motivated by this, we introduce a learning-based surface prediction module (\textbf{SPM}), 
which predicts the coarse complete surface of the target from the current partial reconstruction.
Afterwards, online volumetric mapping extracts incomplete observed surfaces from the prediction and the current reconstruction as the uncovered parts.
Then, a hierarchical planner generates motions for reconstructing the uncovered surfaces in a \textit{coarse-to-fine} manner. It first finds an efficient global path for full coverage.
Secondly, a local path segment from the current pose to the next viewpoint (NBV) is generated under the guidance of the global path while optimizing the crucial factors for MVS performance.
Then, the executable local trajectory is produced to acquire image-pose pairs of the target. 
The collected database is processed by COLMAP \cite{schoenberger2016sfm, schoenberger2016mvs, schoenberger2016vote} for dense 3D reconstruction.

We compare the proposed method with the classical and \textit{state-of-the-art} methods in a realistic simulation. Results present that our method achieves higher efficiency
and better reconstruction quality in benchmark scenarios.
Moreover, benchmark experiments demonstrate the higher autonomy level of our method and our method can realize \textit{real-time}
planning on typical onboard computers.
The contributions of this paper are summarized as follows:

1) A surface prediction module (SPM), which directly infers the complete target surfaces from partial reconstruction information and facilitates efficient global coverage of the target without wasting significant time on extra exploration.

2) A hierarchical planner based on SPM, which sufficiently considers MVS-related factors on the fly and global coverage, achieving higher reconstruction quality and efficiency.

3) Benchmark comparisons that validate the performance of \textbf{PredRecon}. The source code of our implementation has been made public.

\begin{figure}[t]
	\begin{center}        
		\includegraphics[width=0.99\columnwidth]{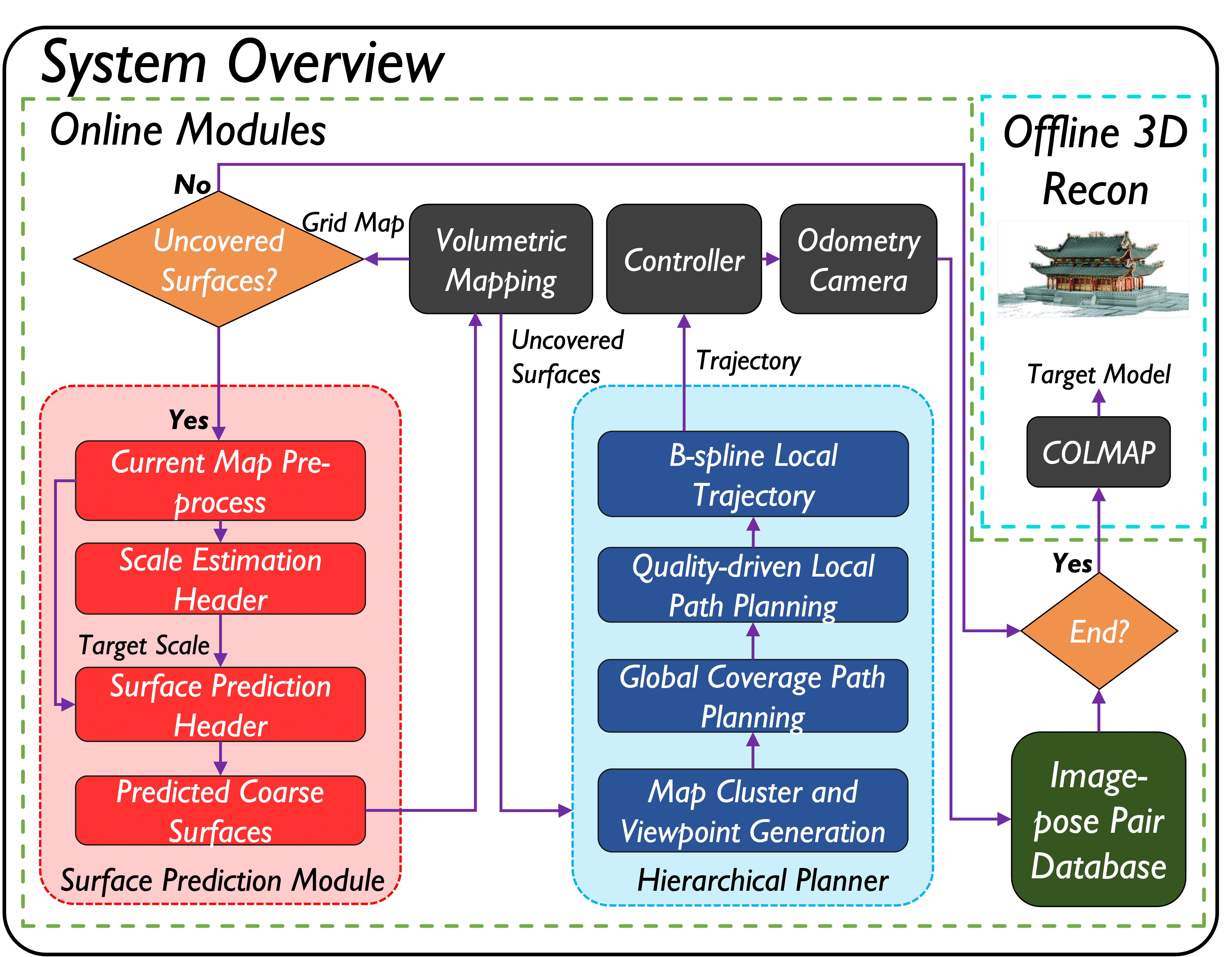}    
      \vspace{-0.7cm}
	\end{center}
   \caption{\label{fig:overview} The overview of the proposed prediction-boosted path planning framework for 3D reconstruction. }
   \vspace{-0.7cm}
\end{figure}

\section{Related Work}
\label{sec:related}

\subsection{Surface Prediction and Completion}

Surface prediction and completion have been an essential topic in 3D reconstruction. 
Existing works can be roughly classified into geometry-based and learning-based methods.

The geometry-based methods predict the entire surface through geometric heuristics from partial input data. Some classical works \cite{kazhdan2013screened, davis2002filling, berger2014state, zhao2007robust} generate complete surface
models using smooth interpolations from incomplete local holes. Those approaches assume that the whole surface can be inferred directly from the geometric input structure. Thus, they cannot work well during most
of the flight time.

The learning-based methods take inputs from point clouds acquired through surface voxelization. They \cite{yuan2018pcn, xie2020grnet, pan2021variational, shi2022temporal} directly output the complete surface model with an implicit parameterized model (deep neural network), 
which has better adaptiveness to complex situations. Our SPM belongs to this category. However, most existing methods suffer from unstable accuracy, primarily influenced by normalization. Hence, an extra detector is essential for predicting the scale and center of the target model. 
Additionally, many apply 3D CNNs for higher accuracy, while heavy architecture leads to slower inference time.

Based on this approach \cite{yuan2018pcn}, our SPM directly uses map point cloud as input and achieves \textit{end-to-end} surface prediction without an extra detector for normalization. Moreover, we optimize the network architecture
with a more lightweight structure and more accurate performance (Sect.\ref{subs:spm_performance}).

\subsection{Path Planning for Aerial Reconstruction}

For efficient and high-quality 3D reconstruction, viewpoints path planning, which selects a minimum quantity of viewpoints while maximizing contributions to reconstruction quality, has been intensely studied for years. The fundamental problem is how to model the bridge 
from viewpoints selection to quality. Several methods \cite{hornung2008image, hepp2018plan3d, vazquez2003automatic} leverage viewpoint information gain (defined as coverage of the coarse model) as the planning objectives. Furthermore, other works \cite{zhang2021continuous, roberts2017submodular}
distribute a coverage hemisphere to each surface, ensuring selected viewpoints scan whole surfaces from diverse view directions. 

MVS-based methods \cite{song2020active, song2021view, smith2018aerial, peng2019adaptive} determine the optimal viewpoints considering MVS factors for better depth estimation, as this paper does.
\cite{song2021view, song2020active} formulate the problem as an \textit{information path planning problem} while \cite{smith2018aerial, peng2019adaptive} adopt a selection strategy based on
\textit{reconstructability} heuristics. They all consider the factor of stereo matching and triangulation.

In this paper, we base our hierarchical planner on MVS-based works but with a more concise formulation of MVS heuristics cost. Moreover, it fully utilizes SPM results to generate paths with high reconstruction efficiency and quality.  

\section{System Overview}
\label{sys}

Fig.\ref{fig:overview} illustrates the overview of the proposed pipeline consisting of online and offline modules. The online subsystem is composed of the SPM (Sect.\ref{sec:spm}), online volumetric mapping (Sect.\ref{subs:pred_states})
and a hierarchical planner (Sect.\ref{sec:hp}). 
SPM predicts both the scale and point cloud of the whole target model surfaces from the current partial map (Sect.\ref{sec:spm}). 
Then, online volumetric mapping extracts the remaining uncovered surfaces with SPM results (Sect.\ref{subs:pred_states}). 
After that, the hierarchical planner works to find a global path and generate a local trajectory for maximizing global coverage efficiency and MVS performance. UAV collects image-pose pairs from odometry and onboard camera (Sect.\ref{sec:hp}). The online subsystem will end the flight if mapping finds no uncovered surfaces.
Afterwards, the image-pose pairs database is processed using offline COLMAP to acquire the 3D reconstruction model of the target.

\begin{figure}[t]
	\begin{center} 
      \vspace{0.5cm}     
      \includegraphics[width=0.99\columnwidth]{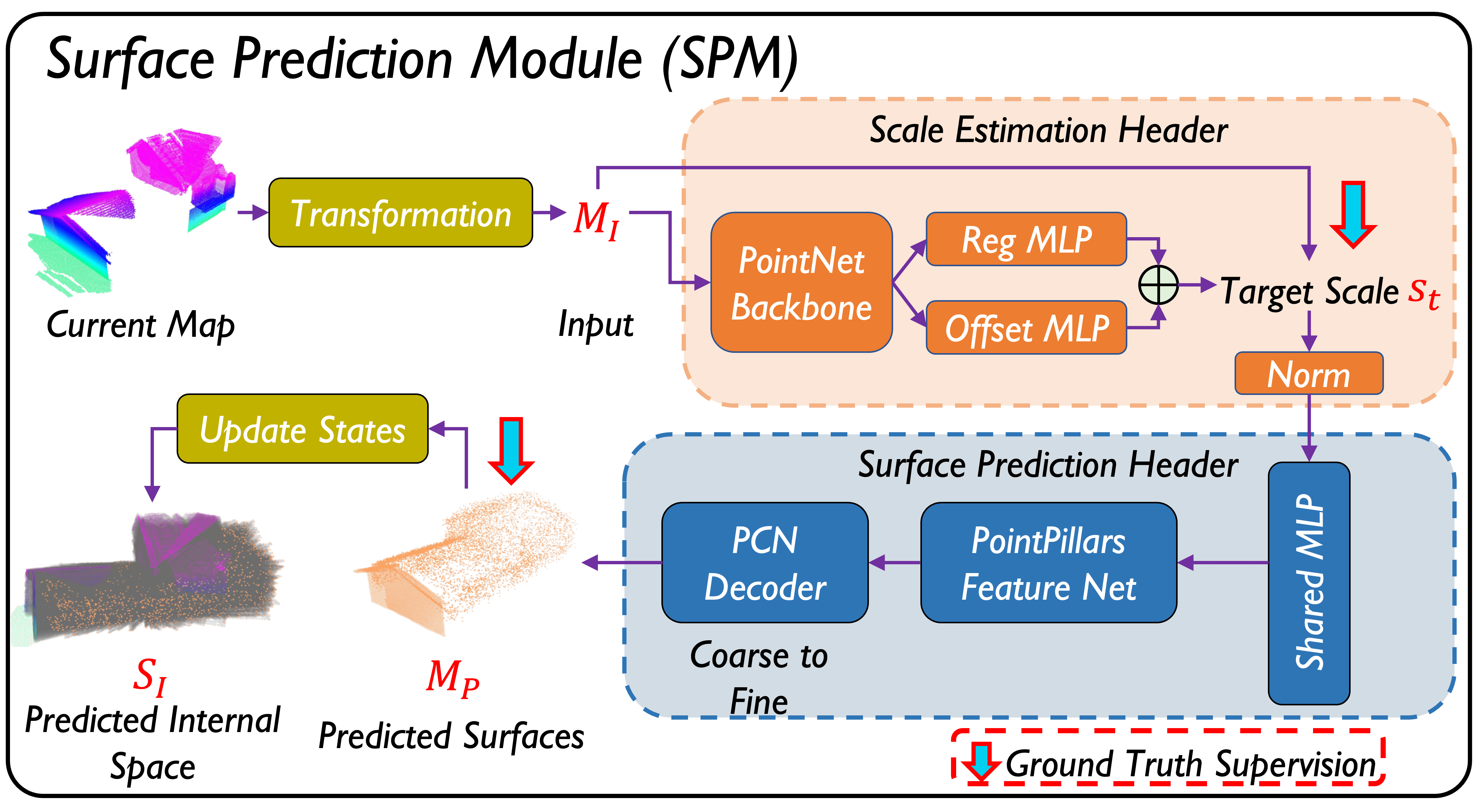} 
      \vspace{-0.7cm}
	\end{center}
   \caption{\label{fig:spm_network} The overall architecture of the proposed SPM (Sect.\ref{sec:spm}).}
   \vspace{-0.8cm}
\end{figure} 

\section{Surface Prediction Module}
\label{sec:spm}

SPM enables predicting the whole surfaces of the target from partial map in entirely unknown environments, as depicted in Fig.\ref{fig:spm_network}.
Surface prediction effectively decreases the redundant flight since no extra time is spent for exploring unknown environments.
Moreover, it facilitates generating fewer viewpoints with the sufficient observation of the target, which reduces the complexity of the subsequent planner.

\subsection{Data Pre-process}
\label{subs:data_pp}
The input of SPM is a down-sampling point cloud $M_{C}$ of the current partial map (Sect.\ref{subs:pred_states}) with the fixed quantity $N_{C}$. Different from previous works \cite{yuan2018pcn, xie2020grnet,pan2021variational}, we directly process each point $p_{i} \in M_{C}$ via a local transformation $T_{p}$, as follows:
\begin{equation}
   T_{p}(p_{i}, C_{C}) = p_{i} - C_{C},
\end{equation}
where $C_{C}$ is the centroid  of $M_{C}$. Then, each transformed point is stored in $M_{I}$, which is sent to the prediction network.

\subsection{Prediction Network Structure}
\label{subs:pred_net}
Compared with previous point cloud completion works \cite{yuan2018pcn, xie2020grnet,pan2021variational,huang2020pf}, our prediction network adopts \textit{end-to-end} manner without the extra detector for normalization. Additionally, it ensures \textit{real-time} and lightweight requirements without 3D convolutional operation in network implementation.
It consists of two headers, the scale estimation header, and the surface prediction header.

To facilitate the following surface prediction, scale estimation header is introduced to predict the coarse scale of the target. The input $M_{I}$ is represented as an $N_{C} \times 3 $ matrix containing the 3D coordinate $(x,y,z)$ of each point. Specifically, we leverage PointNet \cite{qi2017pointnet} as the backbone for its permutation invariance and effective global feature extraction. 
Then, there are two multi-layer perceptrons (MLP) as output branches. Regression MLP directly gives a vector $(x_{s}, y_{s}, z_{s})$ indicating the scales in three axes. To further improve the scale estimation accuracy, the local feature map after PointNet is particularly processed through offset MLP 
to acquire corresponding offset $(\Delta x_{s}, \Delta y_{s}, \Delta z_{s})$. Thus, the target scale $s_{t}$ can be formulated as:
\begin{equation}
   s_{t} = max(x_{s}+\Delta x_{s}, y_{s}+\Delta y_{s}, z_{s}+\Delta z_{s}).
\end{equation}
For the training stage, we use Huber loss to supervise the scale estimations in each axis. Finally, normalization is applied on input point cloud $M_{I}$ by scaling down $s_{t}$-fold.

Surface prediction header is responsible for generating the complete surfaces of the target according to the normalized $M_{I}$. We utilize a shared MLP to encode each point in the normalized $M_{I}$ into the feature map $F$. Then, a PointPillars Feature Net \cite{lang2019pointpillars} is performed on $F$ as the encoder to aggregate geometric information in different areas
with low computation cost for its pseudo image operation. Moreover, PointPillars is eligible for this problem since we expect the network to have the space-aware capability to extend or complete partial surfaces in different areas. Similar to PCN \cite{yuan2018pcn}, a \textit{coarse-to-fine} decoder is also leveraged to generate the prediction for global and local geometry learning.
The fine prediction $Y_{fine}$ and the coarse prediction $Y_{coarse}$ both contain $N_{C}$ points. For the loss function,  the permutation invariant Chamfer Distance is used to supervise the difference between the network outputs with its ground truth $Y_{gt}$, as shown:
\begin{equation}
     cd(X, Y) = \frac{1}{|X|}\sum_{x \in X}\mathop{min}\limits_{y \in Y}||x-y||_{2}^{2}
   +\frac{1}{|Y|}\sum_{y \in Y}\mathop{min}\limits_{x \in X}||x-y||_{2}^{2}
\end{equation}
\begin{equation}
    \mathcal{L} = cd(Y_{coarse}, Y_{gt}) + cd(Y_{fine}, Y_{gt}).
\end{equation}
Afterwards, $M_{I}$ and the inverse normalized $Y_{fine}$ is concatenated into a $2N_{C} \times 3$ matrix as the predicted surfaces $M_{P}$. 
To determine correct viewpoints sampling space, we adopt \textit{GHPR} \cite{katz2015visibility} to process $M_{P}$ to obtain the internal space $S_{I}$, 
which is the prohibited space for viewpoints generation.

\subsection{Volumetric Mapping with Prediction}
\label{subs:pred_states}
To online evaluate the reconstructed parts of the target, we refer to \cite{han2019fiesta} to build a volumetric map, which provides partial observations for SPM. We define the surfaces that are observed from two or more different viewpoints as the complete observed surfaces.
After the inference of SPM, volumetric mapping extracts those incomplete observed surfaces from the prediction as the target uncovered areas of the hierarchical planner.

\begin{figure}[t]
	\begin{center}
      \vspace{0.3cm}
      \includegraphics[width=0.85\columnwidth]{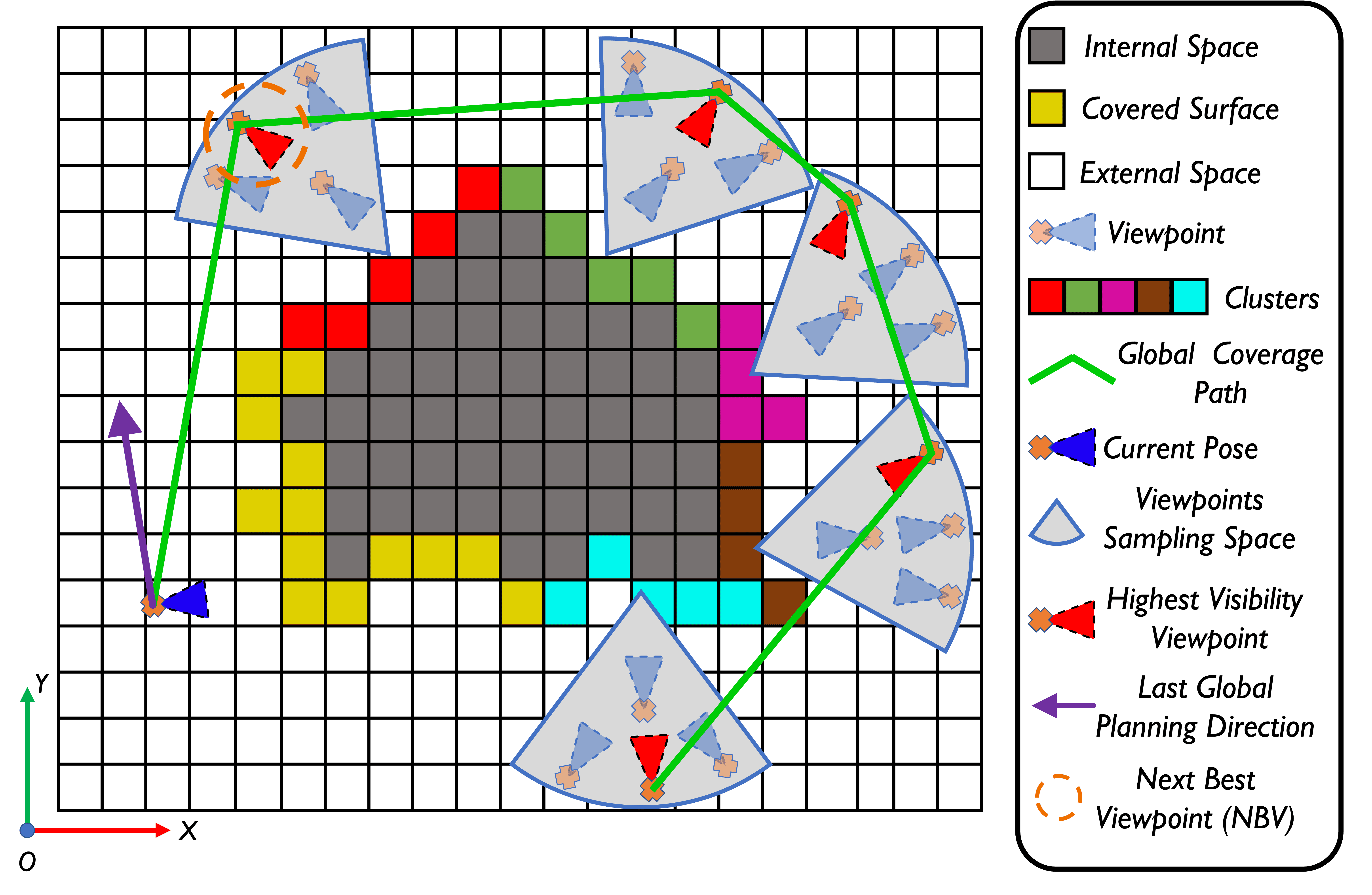}   
      \vspace{-0.5cm}
	\end{center}
   \caption{\label{fig:global} Global coverage path Planning: (1) Cluster the uncovered surfaces. (2) Viewpoints generation through dual sampling. (3) The global coverage path is given by the ATSP solver. (Sect.\ref{subs:global})} 
   \vspace{-0.7cm}
\end{figure}

\section{Hierarchical Planner}
\label{sec:hp}
With the uncovered surfaces, path planning can be formulated as generating paths to efficiently and completely cover the uncovered surfaces of the target.
To realize this objective, the proposed planner takes a hierarchical planning paradigm into two steps, global coverage path planning (Sect.\ref{subs:global}), 
quality-driven local path planning for data collection and trajectory generation (Sect.\ref{subs:local}).

\subsection{Global Coverage Path Planning}
\label{subs:global}
This planning stage is to output an efficient global visit sequence of the viewpoints to cover the uncovered surfaces, as illustrated in Fig.\ref{fig:global}. 
First of all, a clustering approach based on Euclidean distance and normal is performed on the uncovered surfaces to extract $N_{G}$ clusters to be visited. 
Then, similar to \cite{song2021view}, we apply the dual sampling method for the \textit{4-DoF} viewpoints generation, which samples a set of coverage 
viewpoints for each cluster in their own fan-shaped cylinder from its center to normal direction, as shown in Fig.\ref{fig:global}. Lastly, we choose the viewpoint
with the highest surface visibility ratio in each cluster as $V_{G} = \{ v_{g}^{1}, v_{g}^{2}, ..., v_{g}^{N_{G}} \}$, 
where $v_{g}^{i} = (\textbf{\text{P}}_{g}^{i}, \theta_{g}^{i})$ indicating position and yaw angle. The surface visibility ratio of a viewpoint is defined as:
\begin{equation}
   \label{equ:vis_ratio}
   r(v,s) = \frac{\mathcal{N}(v)}{\mathcal{N}(s)},
\end{equation}
where $v$ as viewpoint, $s$ as the observed surface, $\mathcal{N}(v)$ as the number of visible points in $s$ that can be seen from $v$
and $\mathcal{N}(s)$ is the quantity of points in $s$.

To find the shortest path that passes each viewpoint from the current pose, we formulate this problem as the Asymmetric Traveling Salesman Problem (ATSP) \cite{meng2017two}.
The ATSP can be solved by existing proven algorithms through designing proper cost matrix $\Upsilon_{G}$.
Thus, we present the cost between two viewpoints $c_{g}(v_{g}^{i}, v_{g}^{j})$ considers the path length and yaw change, as follows:
\begin{equation}
   \begin{split}
      c_{g}(v_{g}^{i}, v_{g}^{j}) = \frac{L(\textbf{\text{P}}_{g}^{i}, \textbf{\text{P}}_{g}^{j})}{v_{max}} + \\
      \frac{min(||\theta_{g}^{i}-\theta_{g}^{j}||_{1}, 2\pi-||\theta_{g}^{i}-\theta_{g}^{j}||_{1})}{\omega},
   \end{split}
\end{equation}
where $L(\textbf{\text{P}}_{g}^{i}, \textbf{\text{P}}_{g}^{j})$ means the path length between $\textbf{\text{P}}_{g}^{i}$ and $\textbf{\text{P}}_{g}^{j}$ searched by $A^{*}$ algorithm in the free space,
$v_{max}$ and $\omega$ are the maximum velocity and angular change rate of yaw. 

Sometimes, there exist several global coverage paths with similar cost that leads to unstable path optimization results, 
which introduces inconsistent flight directions and low efficiency. Accordingly, global consistency should be essentially taken into account 
to generate stable solutions. We define the last global planning direction (a vector from last current position $\textbf{\text{P}}_{cur}^{last}$ to last NBV $\textbf{\text{P}}_{nbv}^{last}$) $d_{g}^{last}$,  
and introduce glocal consistency cost $c_{GC}(v_{g}^{i})$ by:
\begin{equation}
   d_{g}^{last} = \frac{\textbf{\text{P}}_{nbv}^{last} - \textbf{\text{P}}_{cur}^{last}}{||\textbf{\text{P}}_{nbv}^{last} - \textbf{\text{P}}_{cur}^{last}||_{2}},
\end{equation}
\begin{equation}
   c_{GC}(v_{g}^{i}) = arccos\frac{\textbf{\text{P}}_{g}^{i} - \textbf{\text{P}}_{cur}^{now}}{||\textbf{\text{P}}_{g}^{i} - \textbf{\text{P}}_{cur}^{now}||_{2}} \cdot d_{g}^{last}.
\end{equation}
Then, we can give the complete form of $\Upsilon_{G}$ with the viewpoints index set $ \zeta = \{1,2,...,N_{G}\}$ as:
\begin{equation}
   \label{equ:cost_mat}
   \Upsilon_{G}(k,h) =
   \begin{cases}
      0,&k==h \,or\, h=0\\
      c_{g}(v_{g}^{k}, v_{g}^{h}),&k,h \in \zeta\\
      [\beta_{1}c_{g}(v_{g}^{k}, v_{g}^{h}) + & k==0\,and\,h\in \zeta\\
      \beta_{2}c_{GC}(v_{g}^{h})],
   \end{cases}
\end{equation}
Therefore, through solving the above ATSP with $\Upsilon_{G}$, we can find the efficient global coverage path starting from the current pose to visit the whole uncovered surfaces.

\subsection{Quality-driven Local Path Planning}
\label{subs:local}
\begin{figure}[t]
	\begin{center}
      \includegraphics[width=0.85\columnwidth]{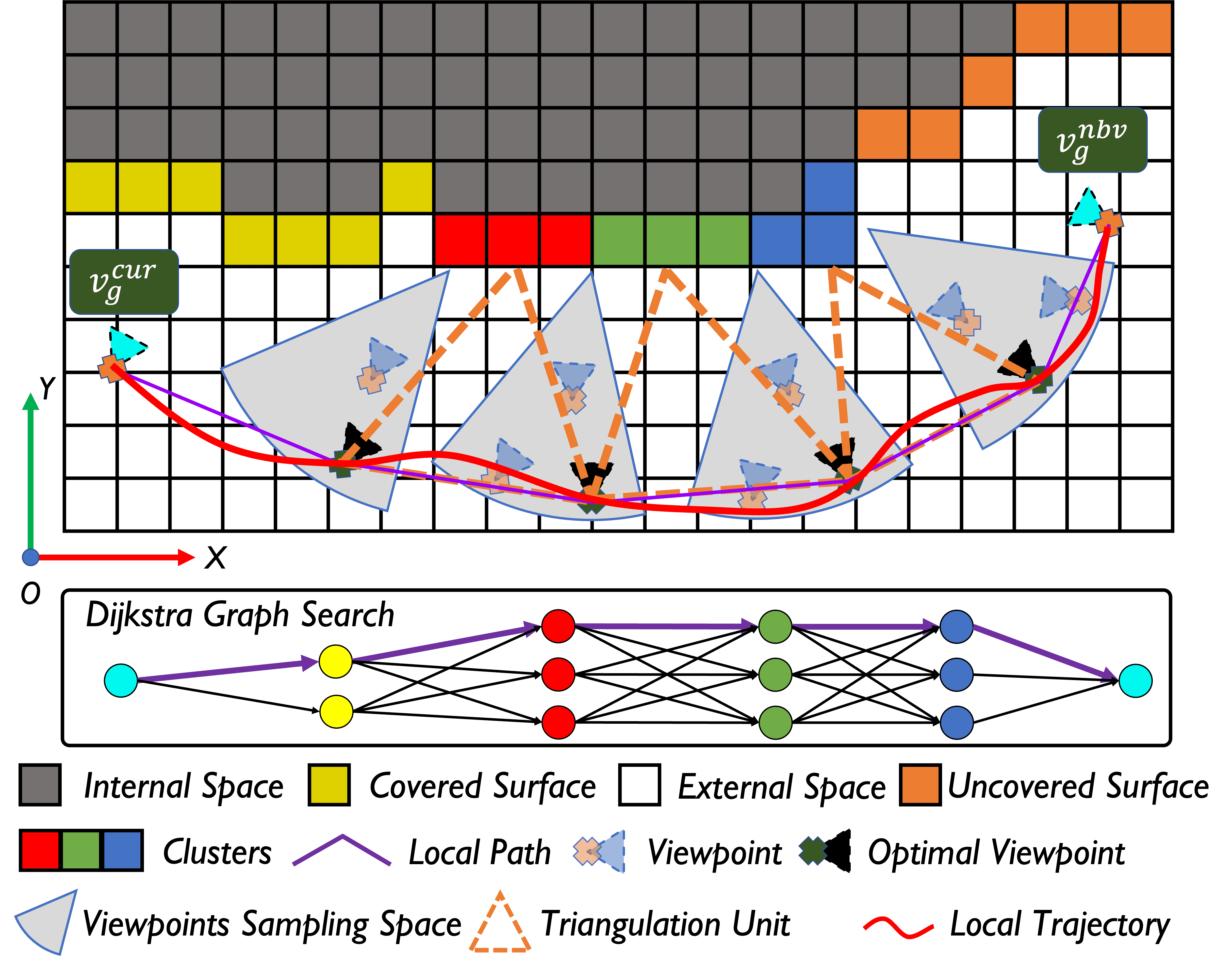}   
      \vspace{-0.5cm}
	\end{center}
   \caption{\label{fig:local} Quality-driven local path planning based on the graph search. 
   Through fully considering MVS-related factors, 
   a reconstruction quality-driven local path is produced with its corresponding trajectory. (Sect.\ref{subs:local})} 
   \vspace{-1.4cm}
\end{figure}

Global planning mainly focuses on fast and complete coverage of the target. To further improve the reconstruction quality,
local planning optimizes a segment path from the current pose to NBV, which fully considers MVS-related factors, as depicted in Fig.\ref{fig:local}.

Different from global planning, the cluster covered by the local segment is further subdivided into smaller clusters while viewpoints sampling space in local planning is determined by two neighboring clusters, as shown in Fig.\ref{fig:local}.
Local viewpoints set is represented as the form of $V_{L} = \{ VP_{1}:\{v_{l}^{1,1}, v_{l}^{1,2}, ..., v_{l}^{1,n} \}, ... , VP_{i}:\{v_{l}^{i,1}, v_{l}^{i,2}, ..., v_{l}^{i,k}, ... \} \}$, and clusters shown as $\mathcal{C}_{L} = \{ cls_{1}, cls_{2}, ..., cls_{j}, ... \}$. 

Many previous studies \cite{schoenberger2016mvs,mendes2016next,mendez2017taking} demonstrated the high-quality MVS reconstruction thoroughly depending on the following factors, including visibility $\mathcal{S}_{vis}$,
relative distance $\mathcal{S}_{dis}$ and triangulation angle $\mathcal{S}_{ang}$, presented in Eq.\ref{equ:quality}, \ref{equ:vis}, \ref{equ:dis}, \ref{equ:ang}. 
To optimize MVS performance of a local path, we decompose the MVS structure into several basic triangulation units, which is defined as each of two neighboring viewpoints in the local path with their co-visible cluster surface.
Furthermore, the MVS performance of this path can be viewed as the reconstruction quality $Q$ sum of all triangulation units in this path.
Then, $Q$ of a triangulation unit can be written as:
\begin{equation}
   \label{equ:quality}
   Q(v_{1}, v_{2}, s) = \mathcal{S}_{vis} \cdot \mathcal{S}_{dis} \cdot \mathcal{S}_{ang},
\end{equation}
where the cluster surface $s$ under two viewpoints $v_{1}$ and $v_{2}$.

$\mathcal{S}_{vis}$ is the score for the visibility ratio ($r \in [0,1]$) of two viewpoints, shown as:
\begin{equation}
   \label{equ:vis}
   \mathcal{S}_{vis}(v_{1}, v_{2}, s) = \frac{r(v_{1}, s) + r(v_{1}, s)}{2}.
\end{equation}
Let $dis_{1}$ and $dis_{2}$ be the distances from two viewpoints to the surface centroid.
We expect $\mathcal{S}_{dis}$ to be close to 1 which leads to similar resolution in two viewpoints images for better depth estimation. The formula follows:
\begin{equation}
   \label{equ:dis}
   \mathcal{S}_{dis}(v_{1}, v_{2}, s) = \frac{min(dis_{1}, dis_{2})}{max(dis_{1}, dis_{2})}.
\end{equation}
$\mathcal{S}_{ang}$ measures the triangulation performance, both accuracy and matchability. 
Let $\epsilon$ be the angle between $vec_{1}$ and $vec_{2}$.
$\varepsilon_{1}$ is the angle between the normal $\mathcal{N}_{s}$ of $s$ and $vec_{1}$ while $\varepsilon_{2}$ is the same for $vec_{2}$. 
Hence, $\mathcal{S}_{ang}$ can be written as:
\begin{equation}
   \label{equ:ang}
      \begin{split}
         vec_{h} & = \mathcal{C}_{s} - v_{h}, \\
         \mathcal{S}_{ang}(v_{1}, v_{2}, s) & = exp(-(\frac{\epsilon-\epsilon_{d}+\varepsilon_{1}-\varepsilon_{2}}{\kappa})^2),
      \end{split}
\end{equation}
where $\mathcal{C}_{s}$ is the centroid of $s$, $\epsilon_{d}$ is the desired triangulation angle and $\kappa$ is a small constant value for numerical stability.

Thus, we can formulate the MVS heuristics cost $c_{MVS}$ and total cost $c_{l}$ with movement cost as:
\begin{equation}
   c_{MVS}(v_{1}, v_{2}, s) = \frac{1}{Q(v_{1}, v_{2}, s)},
\end{equation}
\begin{equation}
   \label{equ:local_cost}
   c_{l}(v_{1}, v_{2}, s) = \alpha_{1} c_{MVS}(v_{1}, v_{2}, s) + (1-\alpha_{1})c_{g}(v_{1}, v_{2}).
\end{equation}
Assuming there are $N_{L}$ clusters totally, the number of $V_{L}$ should be $N_{L}+1$ to satisfy the $N_{L}$ defined triangulation units. 
To optimize the quality-driven cost $c_{l}$ of the local path, we formulate it as a graph search problem. Then, the Dijkstra algorithm is adopted to search for the optimal local path,
$\mathcal{P}_{L} = \{ v_{l}^{1,i_1}, v_{l}^{2,i_2}, ..., v_{l}^{N_{L}+1,i_{N_{L}+1}} \} $ that minimizes the proposed cost:
\begin{equation}
   \mathop{min} \sum_{k=1}^{N_{L}}c_{l}(v_{l}^{k,i_k}, v_{l}^{k+1, i_{k+1}}, cls_{k}).
\end{equation}
Lastly, through leveraging \cite{zhou2019robust}, we convert the local path $\mathcal{P}_{L}$ to the safe, smooth, dynamically feasible, and minimum-time B-spline local trajectory considering
MVS performance to realize an effective collection of image-pose pairs.

\begin{figure*}[t]
	\begin{center}
      \includegraphics[width=1.99\columnwidth]{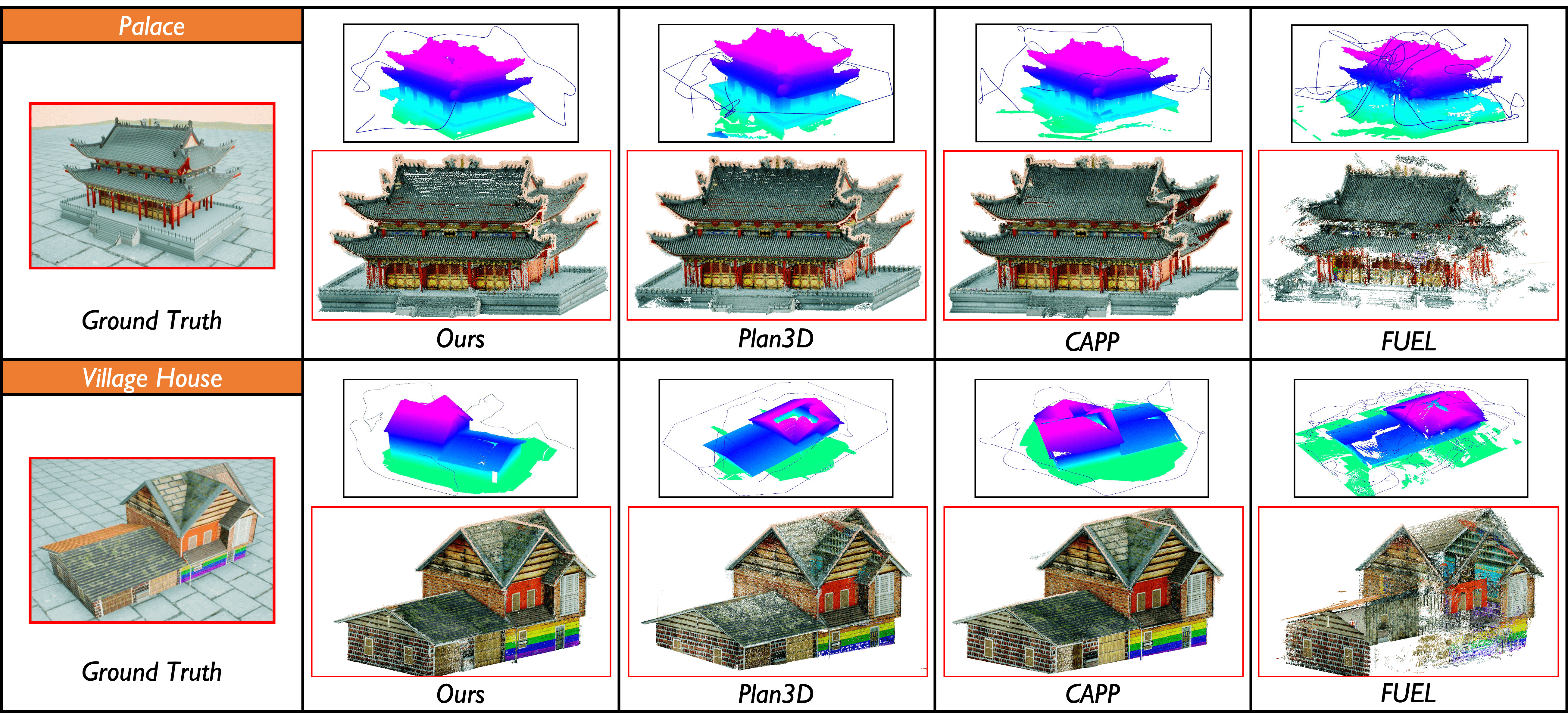}   
      \vspace{-0.5cm}
	\end{center}
   \caption{\label{fig:experiment} Benchmark comparisons (Reconstructed 3D models and volumetric maps with the executed trajectories) of the proposed method, Plan3D \cite{hepp2018plan3d},
   CAPP \cite{zhang2021continuous} and FUEL \cite{zhou2021fuel} in two scenarios (\textbf{\textit{Palace}} and \textbf{\textit{Village House}}).}
   \vspace{-0.4cm}
\end{figure*}

\section{Experiments}
\label{sec:exp}

\subsection{Implementation Details}
\label{subs:imp}
To train our SPM, we use a synthetic CAD model set, Houses3K \cite{peralta2020next} to create a construction scene dataset containing partial and complete
point clouds. Also, we collect other types of construction models in Unreal Engine (UE4\footnote[1]{https://www.unrealengine.com/en-US/}). Specially,
we leverage Blender\footnote[2]{https://www.blender.org/} to generate partial point clouds with 12900 models from different construction categories.
Additionally, we set $N_{C} = 8192$ in the data pre-processing phase. As for training details, the SPM is trained for 200 epochs on single NVIDIA RTX 3070Ti taking 13 hours.
We choose the Adam \cite{kingma2014adam} optimizer during training with an initial learning rate of 1e-4 with a batch size of 16, decaying to 1e-5 at 150 epochs. 

In hierarchical planning, we set $\beta_{1}=1.0$ and $\beta_{2}=5.0$ in Eq.\ref{equ:cost_mat}, $\epsilon_{d}=22.5^{\circ}$ and $\kappa=0.2$ in Eq.\ref{equ:ang},
and $\alpha_{1}=0.8$ in Eq.\ref{equ:local_cost}. In global coverage path planning, the ATSP is solved through a Lin-Kernighan-Helsgaun heuristic solver
\cite{helsgaun2000effective}. 

In all experiments, a geometric controller \cite{lee2010geometric} is used for tracking control of the $(x,y,z,\theta)$ trajectory. SPM runs on an NVIDIA
RTX 3070 Ti (GPU Memory-Usage: \textbf{$\sim$1GB}) and other modules run on an Intel Core i9-10900K CPU.

\subsection{Benchmark Comparisons}
\label{subs:bmk}

We conduct simulation Experiments in a realistic simulator, AirSim in UE4. We benchmark it in two highly textured scenarios,
\textbf{\textit{Palace}} ($15 \times 25 \times 14m^3$) and \textbf{\textit{Village House}} ($14 \times 11 \times 12m^3$). The proposed
method is compared with three methods: Plan3D \cite{hepp2018plan3d} (\textit{explore-then-exploit}), CAPP \cite{zhang2021continuous} (prior-based)
and FUEL \cite{zhou2021fuel} (exploration-based). There is no open source code for Plan3D \cite{hepp2018plan3d} and CAPP \cite{zhang2021continuous},
so we use our implementation. A UAV mounting a forward-looking camera with FOV $[80^{\circ}, 60^{\circ}]$ is adopted as the experimental platform.
It captures images with a resolution $1280 \times 720$ px. In both scenarios, we limit the $v_{max}=0.85m/s$ and $\omega = 0.5 rad/s$.
Plan3D \cite{hepp2018plan3d} firstly executes a pre-defined flight for the coarse model, and then generates the global path using our planner. 
CAPP \cite{zhang2021continuous} produces a global coverage path also by our planner according to input prior model. As for FUEL \cite{zhou2021fuel},
it collects image-pose pairs of the target while exploring the unknown environments containing the target. The collected data of each method is 
processed through COLMAP to obtain reconstructed 3D models.

\begin{table}
   \renewcommand\arraystretch{1.4}
   \tabcolsep=1mm
   \centering
   \caption{Path Planning and 3D Reconstruction results in two scenarios.  \label{tab:benchmark}}
   \begin{tabular}{cccccccc} 
   \hline
                     & Method  & \begin{tabular}[c]{@{}c@{}}Prior \\Model \end{tabular} & \begin{tabular}[c]{@{}c@{}}Path \\Length ($m$) \end{tabular} & \begin{tabular}[c]{@{}c@{}}Time \\($s$)\end{tabular} & \begin{tabular}[c]{@{}c@{}}Recall \\(\%)\end{tabular}  & \begin{tabular}[c]{@{}c@{}}Precision \\(\%)\end{tabular} & \begin{tabular}[c]{@{}c@{}}F-score \\(\%)\end{tabular} \\
   \hline
   \hline
   \multirow{4}{*}{\rotatebox{90}{\textit{Palace}}}     & Plan3D\cite{hepp2018plan3d}      & \XSolidBrush & 375.5                                                         & 507.7                                                        & 74.48 & 82.57 &78.32                                                           \\ 
   \cline{2-8}
                             & CAPP\cite{zhang2021continuous}        & \textcolor{red}{\Checkmark} & 243.6                                                         & 322.6                                                          & 69.21 &85.86 &76.64                                                           \\ 
   \cline{2-8}
                             & FUEL\cite{zhou2021fuel}                    & \XSolidBrush  & 371.1                                                     & 469.8                                                        & 40.31 & 38.38 & 39.32                                                          \\ 
   \cline{2-8}
                             & Ours              & \XSolidBrush        & \textbf{213.1}                                                    & \textbf{252.7}                                                         & \textbf{74.67} & \textbf{86.45} &\textbf{80.13}                                                           \\ 
   \hline
   \multirow{4}{*}{\rotatebox{90}{\textit{Village House}}} & Plan3D\cite{hepp2018plan3d}       & \XSolidBrush  & 239.3                                                         & 310.6                                                         & 64.28 &72.86 &68.30                                                           \\ 
   \cline{2-8}
                             & CAPP\cite{zhang2021continuous}      & \textcolor{red}{\Checkmark}  & 193.4                                                         & 242.3                                                          & 80.30 &\textbf{84.60} &82.40                                                           \\ 
   \cline{2-8}
                             & FUEL\cite{zhou2021fuel}                 & \XSolidBrush    & 405.1                                                       & 506.8                                                         & 44.35 &36.46 &40.02                                                           \\
   \cline{2-8}
                             & Ours                  & \XSolidBrush   & \textbf{153.2}                                                      & \textbf{184.6}                                                         & \textbf{84.54} &83.13 &\textbf{83.83}                                                          \\
   \hline
   \end{tabular}
   \vspace{-0.7cm}
\end{table}

We evaluate their performance by two metrics, efficiency (path length and time) and reconstruction quality (\textit{F-score}). 
The average comparison results are listed in Table.\ref{tab:benchmark} and Fig.\ref{fig:experiment}. Compared with the other methods, we both achieve much shorter time and path length,
primarily since our planner gives a more efficient global coverage path with the support of SPM predictions.
As for reconstruction quality, we refer to the evaluation process and metrics in \cite{knapitsch2017tanks}. First, we perform point cloud alignment between
the reconstructed model and ground truth. Then, two point clouds are uniformly resampled with a voxel size of $0.05m$, which are compared by $Precision$ and $Recall$.
$Precision$ is presented as the percentage of reconstructed points close to a ground truth point while $Recall$ is defined as the percentage of ground truth points
close to a reconstructed point. We set the distance between two points is less than $0.1m$, which are close points. Afterwards, the \textit{F-score} is formulated as 
$F-score = \frac{2(Precision \times Recall)}{Precision+Recall}$.
Fig.\ref{fig:experiment} and Table.\ref{tab:benchmark} depicts the reconstruction quality in two scenarios of each reconstructed model by four methods.
Obviously, the proposed method achieves higher $Precision$, $Recall$, and \textit{F-score}, mainly because our local planning aims to optimize MVS performance,
and our method real-time replans the paths for complete details whenever predictions and map are updated. 
Although our $Precision$ is slightly lower than CAPP \cite{zhang2021continuous} in \textbf{\textit{Village House}} scenario, no prior model is required in our method.

\begin{table}
   \renewcommand\arraystretch{1.4}
   \tabcolsep=1mm
   \centering
   \caption{Computation time of each module. \label{tab:real-time}}
   \begin{tabular}{l|ccccc} 
   \hline
                     & SPM  & \begin{tabular}[c]{@{}c@{}}Global \\Planning \end{tabular} & \begin{tabular}[c]{@{}c@{}}Local \\Planning \end{tabular} & \begin{tabular}[c]{@{}c@{}}Traj. \\Opt.\end{tabular} & \begin{tabular}[c]{@{}c@{}}Total \\Comp.\end{tabular}  \\
   \hline
   \multirow{1}{*}{{Time ($ms$)}}     &$\sim$26.8       &$\sim$93.5                                                          & $\sim$0.5                                                       & $\sim$3.7               & $\sim$124.7                                           \\ 
   \hline
   \end{tabular}
   \vspace{-0.72cm}
\end{table}

As shown in Table.\ref{tab:real-time}, the proposed system can finish planning once in approximately $100ms$, which enables enough frequency for 
\textit{real-time} planning on the onboard computer of a realistic UAV.

\subsection{SPM Prediction Performance}
\label{subs:spm_performance}
Compared with the point cloud completion task, the surface prediction in our system is more difficult since no exact scale and
center are given for normalization. However, under Chamfer Distance and \textit{F-score} metrics, our SPM without prior scale and center still outperforms PCN \cite{yuan2018pcn} in the above task using the generated data (Sect.\ref{subs:imp}) (\textcolor{red}{Left}) and ShapeNet dataset (\textcolor{red}{Right}), as listed in Table.\ref{tab:spm}.
Considering reconstructed surfaces, PCN \cite{yuan2018pcn} produces smoother surfaces than coarse prediction results generated by SPM.

\begin{table}
\scriptsize
   \renewcommand\arraystretch{1.4}
   \tabcolsep=1mm
   \centering
   \caption{Point cloud completion Performance Comparisons. \label{tab:spm}}
   \begin{tabular}{l|cccc} 
   \hline
   Method                  & $\#$Param(M) & L1\_CD (1e-3$m$)  & L2\_CD (1e-4$m$) & F-score (\%) \\
   \hline
   \hline
   our SPM    & \textbf{28.20} &\textbf{13.6404} / \textbf{9.4461}       &\textbf{14.7100} / \textbf{3.9368}                                                           &\textbf{52.6050} / \textbf{68.6693}                                                                                              \\ 
   \hline
   PCN \cite{yuan2018pcn} & 28.91 &15.5221 / 10.4897       &18.3987 / 4.7431                                                          &50.1210 / 65.7207                                                                                                \\
   \hline   
\end{tabular}
   \vspace{-0.72cm}
\end{table}

\section{Conclusions}
\label{sec:conclusion}
In this paper, we propose a prediction-boosted planning framework for efficient high-quality 3D reconstruction with an autonomous single trail.
The proposed SPM predicts complete surfaces from the partial map to provide global information for the path planner. Based on the SPM, a hierarchical
planner sequentially plans motions for 3D reconstruction. It finds efficient global coverage paths, optimizes reconstruction quality-driven local paths to improve
MVS performance, and generates smooth corresponding local trajectories. The method significantly improves reconstruction efficiency and quality via introducing
SPM and considering MVS-related factors. Challenging benchmark in realistic simulation shows the competence of \textbf{PredRecon} compared with the existing classical and \textit{state-of-the-art} methods.

The limitation of our method is insufficient real-world tests as well as the limited generalizability and robustness of SPM. In the future, we plan to further optimize SPM architecture for better data representation and implement more challenging real-world tests.

\bibliography{fc}

\begin{thebibliography}{10}
\providecommand{\url}[1]{#1}
\csname url@samestyle\endcsname
\providecommand{\newblock}{\relax}
\providecommand{\bibinfo}[2]{#2}
\providecommand{\BIBentrySTDinterwordspacing}{\spaceskip=0pt\relax}
\providecommand{\BIBentryALTinterwordstretchfactor}{4}
\providecommand{\BIBentryALTinterwordspacing}{\spaceskip=\fontdimen2\font plus
\BIBentryALTinterwordstretchfactor\fontdimen3\font minus
  \fontdimen4\font\relax}
\providecommand{\BIBforeignlanguage}[2]{{%
\expandafter\ifx\csname l@#1\endcsname\relax
\typeout{** WARNING: IEEEtran.bst: No hyphenation pattern has been}%
\typeout{** loaded for the language `#1'. Using the pattern for}%
\typeout{** the default language instead.}%
\else
\language=\csname l@#1\endcsname
\fi
#2}}
\providecommand{\BIBdecl}{\relax}
\BIBdecl

\bibitem{zhang2021continuous}
H.~Zhang, Y.~Yao, K.~Xie, C.-W. Fu, H.~Zhang, and H.~Huang, ``Continuous aerial
  path planning for 3d urban scene reconstruction.'' \emph{ACM Trans. Graph.},
  vol.~40, no.~6, pp. 225--1, 2021.

\bibitem{hepp2018plan3d}
B.~Hepp, M.~Nie{\ss}ner, and O.~Hilliges, ``Plan3d: Viewpoint and trajectory
  optimization for aerial multi-view stereo reconstruction,'' \emph{ACM
  Transactions on Graphics (TOG)}, vol.~38, no.~1, pp. 1--17, 2018.

\bibitem{kuang2020real}
Q.~Kuang, J.~Wu, J.~Pan, and B.~Zhou, ``Real-time uav path planning for
  autonomous urban scene reconstruction,'' in \emph{2020 IEEE International
  Conference on Robotics and Automation (ICRA)}.\hskip 1em plus 0.5em minus
  0.4em\relax IEEE, 2020, pp. 1156--1162.

\bibitem{zhou2020offsite}
X.~Zhou, K.~Xie, K.~Huang, Y.~Liu, Y.~Zhou, M.~Gong, and H.~Huang, ``Offsite
  aerial path planning for efficient urban scene reconstruction,'' \emph{ACM
  Transactions on Graphics (TOG)}, vol.~39, no.~6, pp. 1--16, 2020.

\bibitem{song2021view}
S.~Song, D.~Kim, and S.~Choi, ``View path planning via online multiview stereo
  for 3-d modeling of large-scale structures,'' \emph{IEEE Transactions on
  Robotics}, vol.~38, no.~1, pp. 372--390, 2021.

\bibitem{zhou2021fuel}
B.~Zhou, Y.~Zhang, X.~Chen, and S.~Shen, ``Fuel: Fast uav exploration using
  incremental frontier structure and hierarchical planning,'' \emph{IEEE
  Robotics and Automation Letters}, vol.~6, no.~2, pp. 779--786, 2021.

\bibitem{song2020active}
S.~Song, D.~Kim, and S.~Jo, ``Active 3d modeling via online multi-view
  stereo,'' in \emph{2020 IEEE International Conference on Robotics and
  Automation (ICRA)}.\hskip 1em plus 0.5em minus 0.4em\relax IEEE, 2020, pp.
  5284--5291.

\bibitem{schoenberger2016sfm}
J.~L. Sch\"{o}nberger and J.-M. Frahm, ``Structure-from-motion revisited,'' in
  \emph{Conference on Computer Vision and Pattern Recognition (CVPR)}, 2016.

\bibitem{schoenberger2016mvs}
J.~L. Sch\"{o}nberger, E.~Zheng, M.~Pollefeys, and J.-M. Frahm, ``Pixelwise
  view selection for unstructured multi-view stereo,'' in \emph{European
  Conference on Computer Vision (ECCV)}, 2016.

\bibitem{schoenberger2016vote}
J.~L. Sch\"{o}nberger, T.~Price, T.~Sattler, J.-M. Frahm, and M.~Pollefeys, ``A
  vote-and-verify strategy for fast spatial verification in image retrieval,''
  in \emph{Asian Conference on Computer Vision (ACCV)}, 2016.

\bibitem{kazhdan2013screened}
M.~Kazhdan and H.~Hoppe, ``Screened poisson surface reconstruction,'' \emph{ACM
  Transactions on Graphics (ToG)}, vol.~32, no.~3, pp. 1--13, 2013.

\bibitem{davis2002filling}
J.~Davis, S.~R. Marschner, M.~Garr, and M.~Levoy, ``Filling holes in complex
  surfaces using volumetric diffusion,'' in \emph{Proceedings. First
  international symposium on 3d data processing visualization and
  transmission}.\hskip 1em plus 0.5em minus 0.4em\relax IEEE, 2002, pp.
  428--441.

\bibitem{berger2014state}
M.~Berger, A.~Tagliasacchi, L.~Seversky, P.~Alliez, J.~Levine, A.~Sharf, and
  C.~Silva, ``State of the art in surface reconstruction from point clouds,''
  \emph{Eurographics 2014-State of the Art Reports}, vol.~1, no.~1, pp.
  161--185, 2014.

\bibitem{zhao2007robust}
W.~Zhao, S.~Gao, and H.~Lin, ``A robust hole-filling algorithm for triangular
  mesh,'' \emph{The Visual Computer}, vol.~23, no.~12, pp. 987--997, 2007.

\bibitem{yuan2018pcn}
W.~Yuan, T.~Khot, D.~Held, C.~Mertz, and M.~Hebert, ``Pcn: Point completion
  network,'' in \emph{2018 International Conference on 3D Vision (3DV)}.\hskip
  1em plus 0.5em minus 0.4em\relax IEEE, 2018, pp. 728--737.

\bibitem{xie2020grnet}
H.~Xie, H.~Yao, S.~Zhou, J.~Mao, S.~Zhang, and W.~Sun, ``Grnet: Gridding
  residual network for dense point cloud completion,'' in \emph{European
  Conference on Computer Vision}.\hskip 1em plus 0.5em minus 0.4em\relax
  Springer, 2020, pp. 365--381.

\bibitem{pan2021variational}
L.~Pan, X.~Chen, Z.~Cai, J.~Zhang, H.~Zhao, S.~Yi, and Z.~Liu, ``Variational
  relational point completion network,'' in \emph{Proceedings of the IEEE/CVF
  conference on computer vision and pattern recognition}, 2021, pp. 8524--8533.

\bibitem{shi2022temporal}
J.~Shi, L.~Xu, P.~Li, X.~Chen, and S.~Shen, ``Temporal point cloud completion
  with pose disturbance,'' \emph{IEEE Robotics and Automation Letters}, vol.~7,
  no.~2, pp. 4165--4172, 2022.

\bibitem{hornung2008image}
A.~Hornung, B.~Zeng, and L.~Kobbelt, ``Image selection for improved multi-view
  stereo,'' in \emph{2008 IEEE Conference on Computer Vision and Pattern
  Recognition}.\hskip 1em plus 0.5em minus 0.4em\relax IEEE, 2008, pp. 1--8.

\bibitem{vazquez2003automatic}
P.-P. V{\'a}zquez, M.~Feixas, M.~Sbert, and W.~Heidrich, ``Automatic view
  selection using viewpoint entropy and its application to image-based
  modelling,'' in \emph{Computer Graphics Forum}, vol.~22, no.~4.\hskip 1em
  plus 0.5em minus 0.4em\relax Wiley Online Library, 2003, pp. 689--700.

\bibitem{roberts2017submodular}
M.~Roberts, D.~Dey, A.~Truong, S.~Sinha, S.~Shah, A.~Kapoor, P.~Hanrahan, and
  N.~Joshi, ``Submodular trajectory optimization for aerial 3d scanning,'' in
  \emph{Proceedings of the IEEE International Conference on Computer Vision},
  2017, pp. 5324--5333.

\bibitem{smith2018aerial}
N.~Smith, N.~Moehrle, M.~Goesele, and W.~Heidrich, ``Aerial path planning for
  urban scene reconstruction: A continuous optimization method and benchmark,''
  2018.

\bibitem{peng2019adaptive}
C.~Peng and V.~Isler, ``Adaptive view planning for aerial 3d reconstruction,''
  in \emph{2019 International Conference on Robotics and Automation
  (ICRA)}.\hskip 1em plus 0.5em minus 0.4em\relax IEEE, 2019, pp. 2981--2987.

\bibitem{huang2020pf}
Z.~Huang, Y.~Yu, J.~Xu, F.~Ni, and X.~Le, ``Pf-net: Point fractal network for
  3d point cloud completion,'' in \emph{Proceedings of the IEEE/CVF conference
  on computer vision and pattern recognition}, 2020, pp. 7662--7670.

\bibitem{qi2017pointnet}
C.~R. Qi, H.~Su, K.~Mo, and L.~J. Guibas, ``Pointnet: Deep learning on point
  sets for 3d classification and segmentation,'' in \emph{Proceedings of the
  IEEE conference on computer vision and pattern recognition}, 2017, pp.
  652--660.

\bibitem{lang2019pointpillars}
A.~H. Lang, S.~Vora, H.~Caesar, L.~Zhou, J.~Yang, and O.~Beijbom,
  ``Pointpillars: Fast encoders for object detection from point clouds,'' in
  \emph{Proceedings of the IEEE/CVF conference on computer vision and pattern
  recognition}, 2019, pp. 12\,697--12\,705.

\bibitem{katz2015visibility}
S.~Katz and A.~Tal, ``On the visibility of point clouds,'' in \emph{Proceedings
  of the IEEE International Conference on Computer Vision}, 2015, pp.
  1350--1358.

\bibitem{han2019fiesta}
L.~Han, F.~Gao, B.~Zhou, and S.~Shen, ``Fiesta: Fast incremental euclidean
  distance fields for online motion planning of aerial robots,'' \emph{arXiv
  preprint arXiv:1903.02144}, 2019.

\bibitem{meng2017two}
Z.~Meng, H.~Qin, Z.~Chen, X.~Chen, H.~Sun, F.~Lin, and M.~H. Ang, ``A two-stage
  optimized next-view planning framework for 3-d unknown environment
  exploration, and structural reconstruction,'' \emph{IEEE Robotics and
  Automation Letters}, vol.~2, no.~3, pp. 1680--1687, 2017.

\bibitem{mendes2016next}
O.~Mendes, S.~Hadfield, N.~Pugeault, and R.~Bowden, ``Next-best stereo:
  Extending next-best view optimisation for collaborative sensors,'' 2016.

\bibitem{mendez2017taking}
O.~Mendez, S.~Hadfield, N.~Pugeault, and R.~Bowden, ``Taking the scenic route
  to 3d: Optimising reconstruction from moving cameras,'' in \emph{Proceedings
  of the IEEE International Conference on Computer Vision}, 2017, pp.
  4677--4685.

\bibitem{zhou2019robust}
B.~Zhou, F.~Gao, L.~Wang, C.~Liu, and S.~Shen, ``Robust and efficient quadrotor
  trajectory generation for fast autonomous flight,'' \emph{IEEE Robotics and
  Automation Letters}, vol.~4, no.~4, pp. 3529--3536, 2019.

\bibitem{peralta2020next}
D.~Peralta, J.~Casimiro, A.~M. Nilles, J.~A. Aguilar, R.~Atienza, and
  R.~Cajote, ``Next-best view policy for 3d reconstruction,'' \emph{arXiv
  preprint arXiv:2008.12664}, 2020.

\bibitem{kingma2014adam}
D.~P. Kingma and J.~Ba, ``Adam: A method for stochastic optimization,''
  \emph{arXiv preprint arXiv:1412.6980}, 2014.

\bibitem{helsgaun2000effective}
K.~Helsgaun, ``An effective implementation of the lin--kernighan traveling
  salesman heuristic,'' \emph{European journal of operational research}, vol.
  126, no.~1, pp. 106--130, 2000.

\bibitem{lee2010geometric}
T.~Lee, M.~Leoky, and N.~H. McClamroch, ``Geometric tracking control of a
  quadrotor uav on se (3),'' in \emph{Decision and Control (CDC), 2010 49th
  IEEE Conference on}, 2010, pp. 5420--5425.

\bibitem{knapitsch2017tanks}
A.~Knapitsch, J.~Park, Q.-Y. Zhou, and V.~Koltun, ``Tanks and temples:
  Benchmarking large-scale scene reconstruction,'' \emph{ACM Transactions on
  Graphics (ToG)}, vol.~36, no.~4, pp. 1--13, 2017.

\end{thebibliography}

\end{document}